\documentclass[conference]{IEEEtran}
\IEEEoverridecommandlockouts
\usepackage{cite}
\usepackage{amsmath,amssymb,amsfonts}
\usepackage{algorithmic}
\usepackage{graphicx}
\usepackage{tikz}
\usepackage{textcomp}
\usepackage{xcolor}
\usepackage{mdframed}
\usepackage{hyperref}
\usepackage{booktabs}
\usepackage{xspace}
\usepackage[utf8]{inputenc}
\usepackage{multirow}
\usepackage[switch]{lineno}
\usepackage{pgf-pie}
\usepackage{pgfplots}
\usepackage{pgf}
\usetikzlibrary{patterns}
\usepackage{enumerate}
\usepackage{color,soul}
\usepackage{subfigure}

\newcommand{\etal}{\textit{et al.}\space}
\newcommand{\tool}{\textsc{CABench}\xspace}
\newcommand{\LLMSolve}{\textit{Prompt-to-Solve}\xspace}
\newcommand{\LLMPipe}{\textit{Prompt-to-Pipeline}\xspace}
\newcommand{\hummanRef}{\textit{Human-designed Reference Solution}\xspace}
\newtheorem{definition}{Definition}

\begin{document}

\title{\tool: Benchmarking Composable AI for Solving Complex Tasks through Composing Ready-to-Use Models
}

\author{\IEEEauthorblockN{Tung-Thuy Pham, Duy-Quan Luong, Minh-Quan Duong, Trung-Hieu Nguyen,\\ Thu-Trang Nguyen$^*$\thanks{*Corresponding author.}, Son Nguyen, and Hieu Dinh Vo}
\IEEEauthorblockA{\textit{Faculty of Information Technology} \\
\textit{University of Engineering and Technology, Vietnam National University, Hanoi, Vietnam} \\ \{25025009, 23020140, 23020138, 21020017, trang.nguyen, sonnguyen, hieuvd\}@vnu.edu.vn} 
}

\maketitle

\begin{abstract}

Composable AI offers a scalable and effective paradigm for tackling complex AI tasks by decomposing them into sub-tasks and solving each sub-task using ready-to-use well-trained models. However, systematically evaluating methods under this setting remains largely unexplored. In this paper, we introduce \tool, the first public benchmark comprising 70 realistic composable AI tasks, along with a curated pool of 700 models across multiple modalities and domains.
We also propose an evaluation framework to enable end-to-end assessment of composable AI solutions. 
To establish initial baselines, we provide human-designed reference solutions and compare their performance with two LLM-based approaches. Our results illustrate the promise of composable AI in addressing complex real-world problems while highlighting the need for methods that can fully unlock its potential by automatically generating effective execution pipelines.

\end{abstract}

\section{Introduction}


In recent years, the rapid progress of Artificial Intelligence (AI) has profoundly influenced a wide range of domains, including computer vision~\cite{voulodimos2018deep}, autonomous driving system (ADS)~\cite{nguyen2024generating}, healthcare~\cite{miotto2018deep}, finance~\cite{ozbayoglu2020deep}, etc. AI models have been integrated into modern systems to perform diverse tasks with remarkable performance. These advancements have unlocked the potential of applying AI to tackle increasingly complex problems in real-world scenarios.

However, many real-world tasks are inherently complicated, requiring multiple steps, modalities, and processing components. For instance, a medical diagnostic system may involve the integration of various components such as image analysis, clinical note understanding, and risk assessment. Building a single monolithic AI model capable of handling such diverse and interdependent functionalities is often infeasible~\cite{liang2024taskmatrix}. \textit{This challenge highlights the need for a modular and flexible approach for solving complex AI tasks in practical settings}. 
%

Furthermore, even when a complex task is decomposed into well-defined sub-modules, the process of developing and training specialized models for each component remains highly resource-intensive. It often requires substantial amounts of data, computational resources, and domain expertise. 
Meanwhile, the AI community has released a large collection of well-trained ready-to-use models on public model hubs such as \textit{Hugging Face}~\footnote{\url{https://huggingface.co}}. These models cover a wide range of modalities and tasks, providing a valuable resource for accelerating AI development. This presents a promising opportunity: \textit{instead of training new models from scratch for every complex task, it is more efficient to decompose the problem into manageable sub-tasks and compose appropriate ready-to-use models to solve them}. Leveraging these existing resources facilitates the construction of AI systems which capable of tackling real-world challenges more effectively. Despite this potential, \textit{the problem of automatically decomposing complex AI tasks and composing ready-to-use well-trained models into executable pipelines remains largely underexplored}.

Addressing this problem poses several challenges. First, effective task decomposition must align with the high-level objectives of the original problem while remaining grounded in the practical capabilities of available well-trained models. Defining sub-tasks without considering the functionality and availability of the existing models can lead to decompositions that are impractical to implement. Second, selecting and orchestrating appropriate models for each sub-task requires a comprehensive understanding of both the functionalities of these models and the inter-dependencies among sub-tasks to ensure effective collaboration. Finally, composing the outputs from different models into a coherent final solution entails addressing format mismatches, managing error propagation across stages, and integrating heterogeneous outputs into a unified and reliable result. 

In this paper, we take an initial step toward addressing these challenges by \textit{formulating the problem of Composable AI (CA)}, which seeks to automatically decompose complex tasks into sub-tasks and compose suitable well-trained models to solve them. To support systematic investigation under this setting, we introduce, \tool, a \textit{carefully designed benchmark} comprising \textbf{70 realistic tasks} where CA is essential, along with a curated model pool of approximately \textbf{700 ready-to-use well-trained models}. The tasks and models in \tool span a wide range of topics and modalities, ensuring diversity and practical relevance.
In addition, we propose \textit{a full-pipeline evaluation framework} that enable end-to-end assessment of CA methods, from input tasks to final outputs.
This work aims to establish a foundation for future research in building scalable and efficient AI systems through the principled reuse of existing resources.



To evaluate the practical feasibility of CA, we conduct several experiments comparing two Large Language Model (LLM)-based strategies against \hummanRef across 70 tasks in \tool. The two LLM-based strategies are: (1) \LLMSolve, where the LLM is prompted to directly solve queries of a given task, and (2) \LLMPipe, where the LLM is prompted to construct a solution pipeline by levering the provided model pool. \LLMSolve, which can exploit the LLM's knowledge to produce direct answers, consistently outperforms \LLMPipe. For instance, \LLMSolve achieves an average accuracy of 0.56, which is double the performance of \LLMPipe. Nonetheless, \LLMSolve still falls short of \hummanRef with performance gaps reaching up to 90\%. 
\textit{These results show that while LLMs can address many tasks to some extent, specialized models orchestrated through carefully designed pipelines remain significantly more effective, particularly for tasks requiring compositional reasoning and structured execution.}

The contributions of this paper are as follows:
\begin{enumerate}
    \item \textit{CA Problem Formulation}: We formally define the problem of Composable AI (CA).
    \item \tool: We introduce \tool, the first public benchmark comprising 70 CA tasks, along with a curated model pool of 700 ready-to-use well-trained models.
    \item \textit{Evaluation Framework}: We propose a full-pipeline evaluation framework for systematical end-to-end assessment of CA methods.
    \item \textit{Human-Designed Reference Solutions}: We design and evaluate a manual CA workflow to illustrate practical strategies for task decomposition and model composition.
    \item \textit{Baselines}: We conduct several experiments which automatically solve the tasks in the benchmark and provide the results as baselines for future research.
\end{enumerate}

\noindent Our benchmark and evaluation framework can be found at:
\begin{mdframed}[backgroundcolor=gray!20, linecolor=black]
\href{https://github.com/iSE-UET-VNU/CABench}{https://github.com/iSE-UET-VNU/CABench}
\end{mdframed}

\section{Composable AI Problem Formulation}

\textit{Composable AI (CA)} offers a scalable and efficient paradigm for solving complex, real-world AI tasks by decomposing them into sub-tasks and leveraging suitable available models to addresses each sub-task. To enable systematic study and principled evaluation under this setting, this paper formally defines the problem of CA as follows:

\begin{definition}{\textbf{Composable AI Problem}}.
Given: 
\begin{itemize}
    \item An \textit{AI task} $\mathcal{T}$, characterized by its input space $\mathcal{X}$, output space $\mathcal{Y}$, and objective function $\mathcal{O}$; and
    \item A \textit{pool of ready-to-use models} $\mathcal{M} = \{M_1, \dots, M_n\}$, where each model is specialized by its input/output modalities and functionalities
\end{itemize}
The goal of composable AI is to: 
\begin{enumerate}[(i)]
    \item \textit{\textbf{Decompose}} $\mathcal{T}$ into a set of sub-tasks $\{t_1 \dots, t_k\}$, where each sub-task $t_j$ $(1 \leq j \leq k)$ is associated with a well-defined objective $o_j$ that contributes to the overall objective $\mathcal{O}$;
    \item \textit{\textbf{Select}} appropriate models from the pool $\mathcal{M}$ to solve each sub-task; and
    \item \textit{\textbf{Compose}} the selected models into a \textit{coherent executable pipeline} that, given an input $x \in \mathcal{X}$, produces an output $y \in \mathcal{Y}$ that effectively solves the original task $\mathcal{T}$. 
\end{enumerate}
\end{definition}

Following the above formulation, CA solves a complex task by constructing an executable pipeline that connects selected models, each responsible for a decomposed sub-task.  However, in practice, the models may exhibit incompatibilities in their input/output modalities or data formats, making direct integration difficult. To address these interoperability challenges and enable seamless orchestration, \textit{glue code} is often necessary in the pipeline. 
Glue code refers to (lightweight) modules that perform operations such as data pre/post-processing, format transformation, and the integration of heterogeneous outputs. It ensures consistent data modality and interface compatibility across models, allowing independently developed models to function cohesively within the same pipeline.
Formally, we define a CA solution as follows:

\begin{definition}{\textbf{Composable AI Solution}}.
\label{def:pipeline}
A composable AI solution is an executable pipeline  defined as a directed acyclic graph (DAG) $\mathcal{G} = (\mathcal{V}, \mathcal{E})$, where:
\begin{itemize}
    \item Each node $v \in \mathcal{V}$ represents a computational unit which can be either a selected model invocation or a glue code module; and
    \item Each edge $e \in \mathcal{E}$ $(\mathcal{E} \subseteq \mathcal{V} \times \mathcal{V})$ represents the data flow between computational units, ensuring the output of one node is correctly routed as input to another. 
\end{itemize}
    
\end{definition}

\begin{figure}
    \centering
    \includegraphics[width=\linewidth]{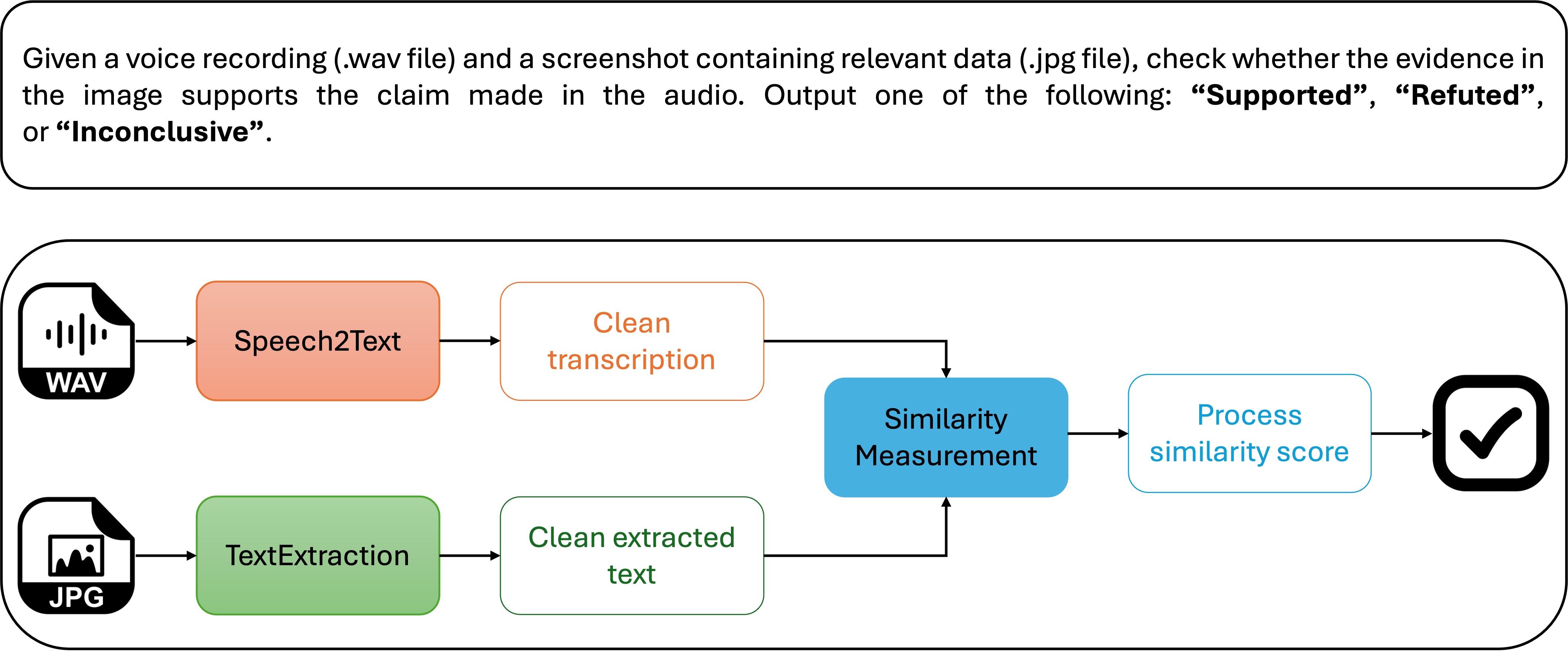}
    \caption{A CA task and its corresponding solution}
    \label{fig:pipeline_example}
\end{figure}

Fig.~\ref{fig:pipeline_example} illustrates an example of a composable AI task and its corresponding solution. 
Given an audio file and an image file, the task is to determine whether the claim made in the audio is \textit{supported}, \textit{refuted}, or \textit{inconclusive} based on textual evidence in the image. To solve this task using the available models from the pool, the proposed solution leverages three models: (1) a \textit{Speech2Text} model (\texttt{openai/whisper-large-v3}) to transcribe the spoken claim into text, (2) a \textit{TextExtraction} model (\texttt{microsoft/trocr-base-printed}) to extract text from the image using OCR, and (3) a \textit{SimilarityMeasurement} model (\texttt{BAII/bge-m3}) to compute the semantic similarity between the transcribed claim and the extracted evidence. 
To enable accurate similarity measurement, both the transcribed audio and the OCR-extracted text should be cleaned, e.g., by standardizing formats and removing noise. These preprocessing operations are performed by dedicated glue code modules.
Additionally, the similarity measurement model \texttt{BAAI/bge-m3} outputs a numerical similarity score, which must be interpreted and mapped to one of the final verdicts: \textit{supported}, \textit{refuted}, or \textit{inconclusive}. This interpretation step is also handled by a glue code snippet.
In summary, the solution consists of six computational units: three model invocations and three glue code modules, collectively forming an executable pipeline to solve the given complex task.

\section{CABench: A Composable AI Benchmark}
\label{sec:benchmark}

To enable systematic investigation and principled evaluation of composable AI, we introduce \tool, a  benchmark specifically designed to reflect the challenges and opportunities of building end-to-end compositional AI solutions. 
Unlike prior benchmarks such as HuggingGPT~\cite{shen2023hugginggpt},
which primarily focus on evaluating the correctness of task decomposition and model selection, \tool goes further by assessing whether the selected models can be correctly composed and executed as a \textit{coherent pipeline} to ultimately fulfill the original objective of the given task.
To the best of our knowledge, \tool is the first public benchmark that supports the evaluation of fully executable CA solutions.

The construction of \tool is guided by several key principles to ensure its utility and relevance for advancing CA. First, it prioritizes \textit{\textbf{realism}} by designing tasks that reflect practical problems grounded in real-world application scenarios. 
Second, it ensures \textit{\textbf{decomposability}} by formulating tasks that can be naturally broken down into meaningful sub-tasks aligned with the overall objective. 
Third, it guarantees \textit{\textbf{solvability}} by ensuring each task can be solved using appropriate models from a defined model pool, with glue code introduced when necessary to enable seamless integration. 
Fourth, it promotes \textit{\textbf{diversity}} by covering a wide range of domains and modalities, ensuring generalization across application areas. 
Finally, it supports \textit{\textbf{evaluability}} by equipping each task with  well-defined input-output specifications, ground-truth outputs, and standardized evaluation metrics, enabling objective, reproducible, and execution-based assessment of CA methods. 


\subsection{Benchmark Construction}
\subsubsection{Task Construction}

Following the above mentioned principles, we construct a benchmark of composable AI tasks designed to reflect the complexity and practical demands of real-world applications. To guarantee \textit{task realism}, each task $\mathcal{T}$ is carefully derived from a popular dataset on Hugging Face\footnote{https://huggingface.co/datasets} and Kaggle\footnote{https://www.kaggle.com/datasets}. Dataset popularity is quantified based on download counts with a threshold of at least 10,000 downloads to ensure broad relevance and practical applicability.
For each selected dataset, we prompt GPT-4o to generate a plausible user story or real-world application scenario that aligns with the dataset's content and structure. These generated descriptions are then manually reviewed, refined, and validated to ensure clarity, feasibility, and alignment with realistic use cases. Through this process, we construct a total of \textit{70 realistic composable AI tasks}, each grounded a widely-used dataset and framed within meaningful application contexts.

Formally, each CA task $\mathcal{T}$ in \tool is represented as $\mathcal{T} = (\mathcal{D}, \mathcal{Q})$ where $\mathcal{D}$ is the \textit{task description}, including an application scenario/user-story, a specification of input space $\mathcal{X}$, output space $\mathcal{Y}$, and an objective function $\mathcal{O}$ that defines the intended transformation from inputs to outputs. Meanwhile, $\mathcal{Q} = \{(x_i, y_i)\}_{i=1}^{N}$ ($x_i\in \mathcal{X},\space y_i \in \mathcal{Y}$) denotes the \textit{query set} consisting of $N$ test with their corresponding ground-truth outputs. This set is partitioned into a validation set $\mathcal{Q}_V$ and a test set $\mathcal{Q}_T$, used respectively for validating and evaluating CA solutions.  The queries in $\mathcal{Q}$ are randomly selected from the original dataset from which the task is derived.

\subsubsection{Model Pool Construction}
To solve the tasks in \tool, one would typically decompose each task into sub-tasks and select appropriate models from open ML model repositories such as Hugging Face, Kaggle, or the ONNX Model Zoo. However, these repositories are large, continuously evolving, and lack standardized configurations, making them unsuitable for consistent and reproducible evaluation in a benchmarking setting. For ease of comparison, we construct a fixed model pool tailored for \tool.

The model pool is curated according to several constraints to ensure quality, diversity, and practical usability. Specifically, we prioritize models that are widely adopted and well-documented on Hugging Face model hub, using these factors as signals of quality. To maintain diversity, we select models with different architectures, trained on varied datasets across multiple domains. Moreover, to accommodate hardware limitations and realistic deployment settings, we restrict the model size to a maximum of 500M parameters. 

\begin{figure}[]
    \centering
    \begin{minipage}[b]{0.48\linewidth}

        \includegraphics[width=0.9\linewidth]{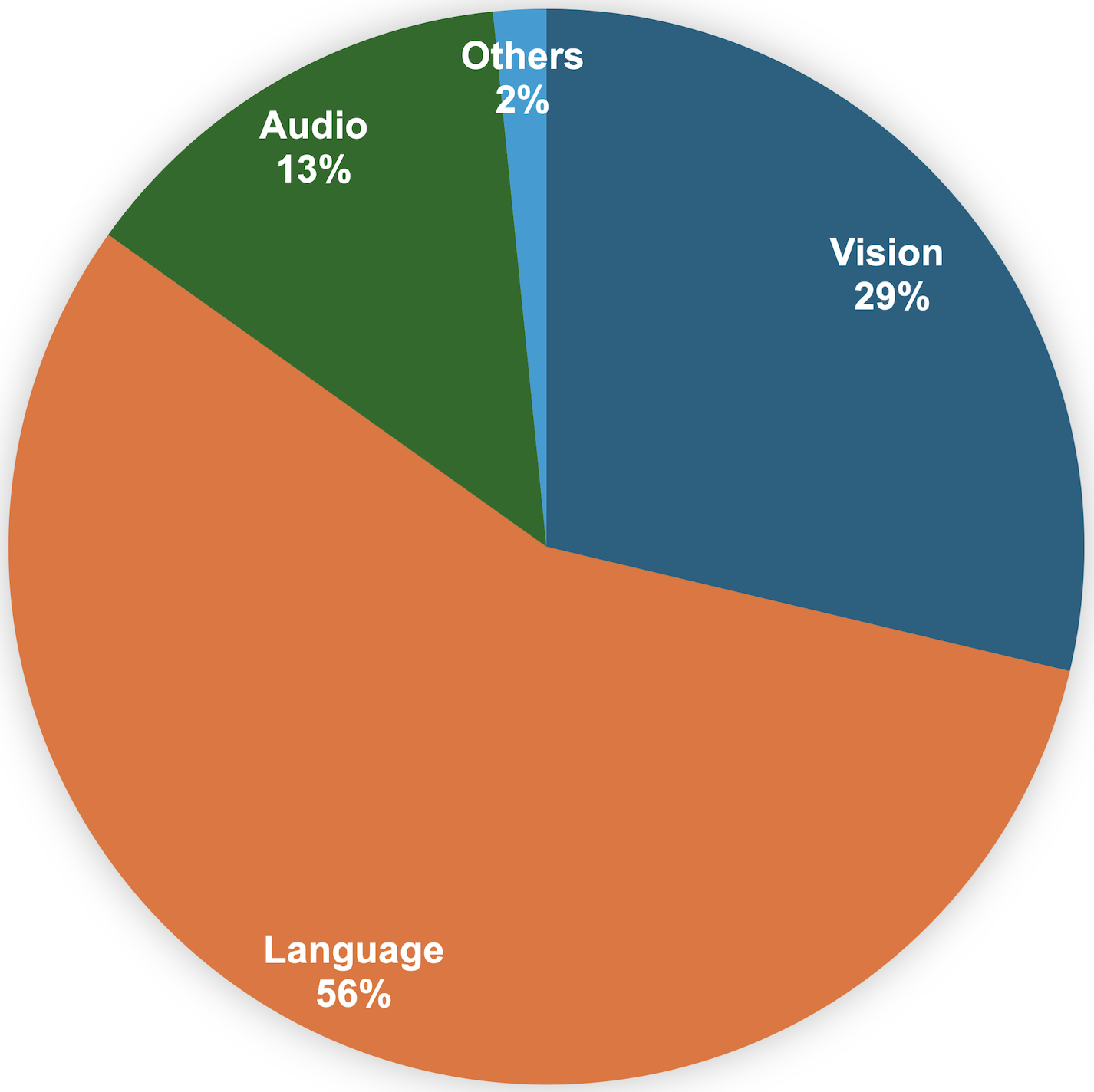}

    \end{minipage}
    \hfill
     \begin{minipage}[b]{0.48\linewidth}
    
        \includegraphics[width=0.9\linewidth]{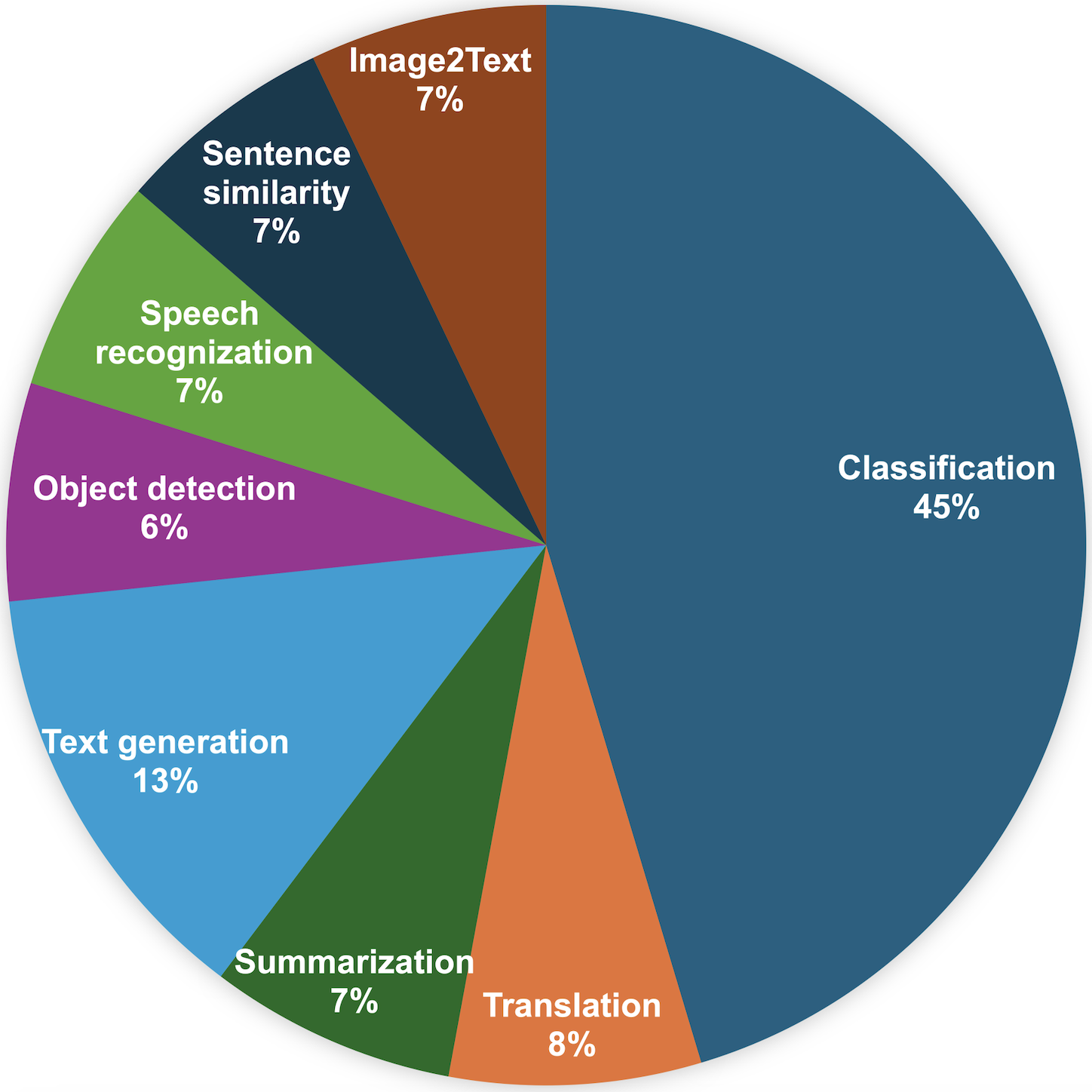}
    
    \end{minipage}
    \caption{ML models in \tool's model pool by modalities and types}
    \label{fig:model_distribution}
\end{figure}

Overall, our model pool consists of about 700 ready-to-use models covering a wide spectrum of modalities and task types. 
Fig.~\ref{fig:model_distribution} presents the distribution of models across modalities and types within the model pool of \tool.
This curated pool ensures that every task in \tool is solvable using the available models while allowing meaningful evaluation of composition strategies under practical constraints.

\subsection{Human-designed Reference Solutions}

To demonstrate that each task in \tool is decomposable and solvable using the curated model pool, we construct reference solutions for all benchmark tasks. Each task is independently addressed by the first four authors, who manually design decomposition strategies, select appropriate models from the pool, and implement necessary glue code. The initial solutions are then collaboratively validated and refined to ensure that the decomposition aligns with the task's objective and that the composed pipeline yields correct outputs. 
Each reference solution is implemented as an execution pipeline, as defined in Definition~\ref{def:pipeline}. 

To facilitate systematic analysis, we categorize the solutions based on their structural complexity. Specifically, solutions are grouped into three categories: \textit{Atomic Solution}, \textit{Chain Solution}, and \textit{Graph Solution}.

\begin{definition}{\textbf{Atomic Solution}}.
\label{def:atomic}
A solution is an atomic solution if its execution pipeline, represented by a DAG $\mathcal{G} = (\mathcal{V}, \mathcal{E})$, consist of a \textit{single node} with no edges, $|\mathcal{V}| = 1$ and $\mathcal{E} = 0$.
    
\end{definition}

\begin{definition}{\textbf{Chain Solution}}.
\label{def:chain}
A solution is a chain solution if its execution pipeline forms a \textit{linear chain}, satisfying $|\mathcal{V}| = k \geq 2$, $|\mathcal{E}| = k-1$, and there exists an ordering of the nodes $\mathcal{V} = \{v_1, \dots, v_k\}$ such that: $\mathcal{E} = \{(v_i, v_{i+1}) | 1 \leq i < k\}$
\end{definition}

\begin{definition}{\textbf{Graph Solution}}.
\label{def:graph}
A solution is a graph solution if its execution pipeline forms a \textit{general DAG} with at least one node having multiple incoming/outgoing edges; formally, $|\mathcal{V}| \geq 2$ and $\exists v \in \mathcal{V}$ such that $out(v) > 1$ or $in(v) > 1$, where:
\begin{itemize}
    \item $out(v) = |\{u \in \mathcal{V}|(v, u) \in \mathcal{E}\}|$ denotes the out-degree of node $v$,
    \item $in(v) = |\{u \in \mathcal{V}\}| (u, v) \in \mathcal{E}|$ denotes the in-degree of node $v$.
\end{itemize}
\end{definition}

\begin{figure}
    \centering
    \includegraphics[width=0.45\linewidth]{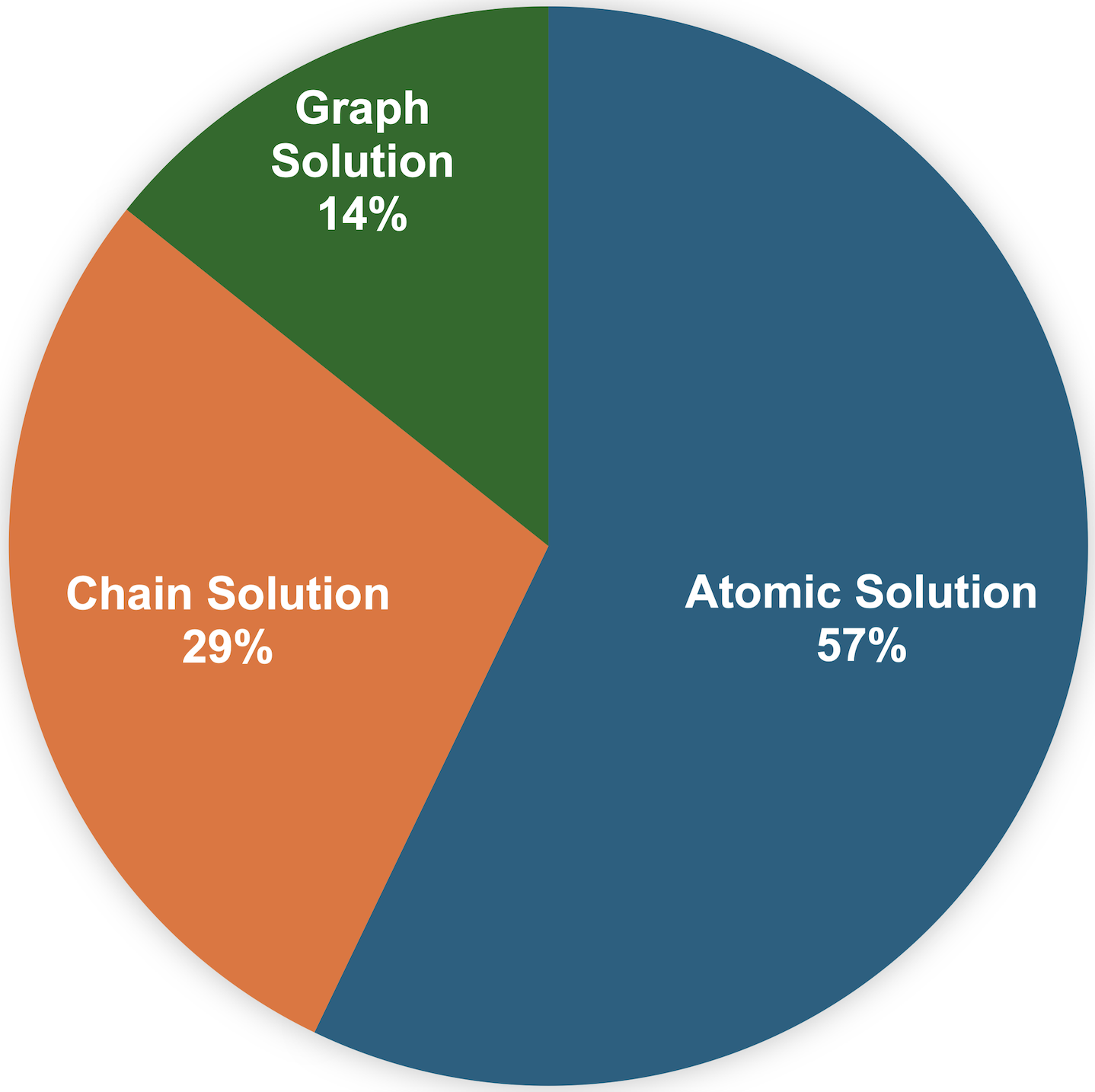}
    \caption{Categorical distribution of human-designed reference solutions}
    \label{fig:solution_category}
\end{figure}



    


Fig.~\ref{fig:solution_category} shows  the distribution of human-designed reference solutions across the three structural categories. It is important to note that a task may admit multiple valid solutions with different pipeline structures depending on the chosen decomposition strategies and model selections.

\subsection{Evaluation Framework}

In addition to the benchmark, we introduce an evaluation framework for automatically assessing the quality of CA solutions within \tool. The framework takes as input a task $\mathcal{T} = (\mathcal{D}, \mathcal{Q})$ and a corresponding solution $\mathcal{G}$, which may be either manually designed or automatically generated by a CA technique. The framework then executes $\mathcal{G}$ to perform inference on the provided query set $\mathcal{Q}$, producing predicted outputs for each query instance. 

To evaluate the performance of  $\mathcal{G}$ in solving  $\mathcal{T}$, the predicted outputs are compared against the ground-truth outputs in $\mathcal{Q}$ using task-appropriate standard metrics\footnote{A complete list of metrics used for each task can be found on our website~\cite{website}.}. For example, Accuracy and F1-Score are used for classification tasks, $R^2$-Score for regression tasks, and BLEU and ROUGE-L for generation tasks. Moreover, to enable consistent comparisons across tasks and methods, all metric scores are linearly normalized to the range $[0,1]$, with higher scores indicating better performance. This normalization allows for cross-task aggregation while preserving relative performance rankings. Overall, this evaluation framework supports objective and automated assessment of CA solutions in \tool, providing a solid foundation for benchmarking future methods.

\section{Experimental Results}

\subsection{Baselines}
In this work, we leverage the capabilities of LLMs to automatically solve the benchmarking tasks in \tool, providing initial baselines alongside \hummanRef. We evaluate the LLM under two strategies: \LLMSolve and \LLMPipe.  Given a task $\mathcal{T}$ with description $\mathcal{D}$ and test set $\mathcal{Q}_T$: 
\begin{itemize}
    \item \textbf{\LLMSolve} directly queries the LLM with the task description $\mathcal{D}$ and each query instance in  $\mathcal{Q}_T$. The LLM is expected to \textit{directly generate final output} for each query. 
    
    \item \textbf{\LLMPipe} instructs the LLM to approach the task in a \textit{composable AI fashion} by explicitly prompting it to (i) decompose the task into sub-tasks, (ii) select suitable models from the provided model pool, and (iii) compose these models and glue code. The LLM is expected to \textit{generate a complete executable pipeline}. This pipeline is then executed to produce the final outputs for the queries in $\mathcal{Q}_T$.
\end{itemize}

For both strategies, we experiment with three prompting patterns, including \textit{Zero-shot}, \textit{Few-shot}, and \textit{Chain-of-Thought (CoT)}~\cite{stechly2024chain}. The prompt templates used for each strategy and prompting pattern are available on our website~\cite{website}.

\subsection{Experimental Procedure and Setup}
In this work, all our experiments are conducted using \text{GPT-4o-mini} with fixed decoding parameters. To reduce the randomness, the temperature is set to $0$. Experiments are conducted on machines equipped with 8-core CPUs, 32GB RAM, and NVIDIA T4 GPUs. Due to the prompt length limitation (130,000 tokens), it is infeasible to include the full descriptions of all models in the entire model pool as context for the \textit{Prompt-to-Pipeline} approach. To address this, we truncate the context appropriately. In these cases, we randomly shuffle the order of the model pool, repeat the experiments five times, and report the averaged results to reduce potential bias introduced by model ordering.

\subsection{Experimental Results}

\subsubsection{Performance of Baselines Across Task Types}

\begin{table*}[]
\caption{Performance of Baselines Across Task Types}
\label{tab:task_type_performance}
\resizebox{\textwidth}{!}{%
\begin{tabular}{ll|c|c|c|c|c|c|c|c|c|c}
\toprule
\multicolumn{2}{c|}{\textbf{Approach}}                                        & \multicolumn{1}{l|}{\begin{tabular}[c]{@{}c@{}}\textbf{Classification}\end{tabular}} & \multicolumn{1}{l|}{\begin{tabular}[c]{@{}c@{}}\textbf{Regression}\end{tabular}} & \multicolumn{1}{l|}{\begin{tabular}[c]{@{}c@{}}\textbf{Question} \\ \textbf{Answering}\end{tabular}} & \multicolumn{1}{l|}{\begin{tabular}[c]{@{}c@{}}\textbf{Summarization}\end{tabular}} & \multicolumn{1}{l|}{\begin{tabular}[c]{@{}c@{}}\textbf{Translation}\end{tabular}} & \multicolumn{1}{l|}{\begin{tabular}[c]{@{}c@{}}\textbf{Object Detection}\end{tabular}} & \multicolumn{1}{l|}{\begin{tabular}[c]{@{}c@{}}\textbf{Generation}\end{tabular}} & \multicolumn{1}{l|}{\begin{tabular}[c]{@{}c@{}}\textbf{Image2Text}\end{tabular}} & \multicolumn{1}{l|}{\begin{tabular}[c]{@{}c@{}}\textbf{Speech} \\ \textbf{Recognition}\end{tabular}} & \multicolumn{1}{l}{\begin{tabular}[c]{@{}c@{}}\textbf{Sentence} \\ \textbf{Similarity}\end{tabular}} \\ 
\midrule
\multicolumn{1}{l|}{\multirow{3}{*}{Prompt-to-Solve}}    & Zero-shot & 0.54 & 0.15 & 0.51 & 0.52 & 0.74 & 0.48 & 0.57 & 0.45 & 0.00 & 0.00                         \\  
\multicolumn{1}{l|}{}                                    & Few-shot  & 0.55 & 0.15 & 0.51 & 0.53 & 0.75 & 0.49 & 0.55 & 0.45 & 0.00 & 0.00                          \\ 
\multicolumn{1}{l|}{}                                    & CoT       & 0.54 & 0.15 & 0.47 & 0.51 & 0.74 & 0.44 & 0.50 & 0.45 & 0.00 & 0.00                         \\ \midrule
\multicolumn{1}{l|}{\multirow{3}{*}{Prompt-to-Pipeline}} & Zero-shot & 0.09 & 0.00 & 0.00 & 0.06 & 0.03 & 0.16 & 0.23 & 0.05 & 0.03 & 0.05                         \\  
\multicolumn{1}{l|}{}                                    & Few-shot  & 0.17 & 0.00 & 0.00 & 0.06 & 0.40 & 0.32 & 0.25 & 0.05 & 0.07 & 0.10                        \\ 
\multicolumn{1}{l|}{}                                    & CoT       & 0.23 & 0.00 & 0.00 & 0.05 & 0.23 & 0.38 & 0.25 & 0.06 & 0.02 & 0.03                       \\ \midrule
\multicolumn{2}{l|}{Human-designed reference}                        & 0.80 & 0.85 & 0.85 & 0.49 & 0.76 & 0.83 & 0.29 & 0.67 & 0.95 & 0.95                         \\ \bottomrule
\end{tabular}
}
\end{table*}

The performance of two automated approaches, i.e., \LLMSolve and \LLMPipe,  and the manual approach, \hummanRef across different task types is shown in Table~\ref{tab:task_type_performance}. For each task, the performance of the approaches is measured using the appropriate standard metrics and then linearly normalized to the range $[0, 1]$, where higher values indicates better performance.

For almost all task types, \textit{\LLMSolve demonstrates superior performance over \LLMPipe}. For example, in translation tasks, \LLMSolve achieves an average score of 0.74, which is more than three times higher than that of \LLMPipe. Notably, \LLMPipe fails to solve several task types such as regression and question answering, i.e., yielding scores of zero, whereas \LLMSolve successfully handles these tasks, achieving normalized scores ranging from 0.15 to 0.50.

These results suggest that LLMs are more effective at generating direct answers than orchestrating multi-model pipelines from scratch.
The advantage arises from \LLMSolve's ability to  leverage the LLM's extensive pre-trained knowledge to directly produce answers for given input queries, allowing it to effectively handle many tasks without explicitly decomposition. 
Meanwhile, \LLMPipe requires the LLM to not only understand the task structure but also select appropriate models and compose them into a valid execution pipeline, which is an inherently challenging process.
Moreover, the prompt length limitation also prevents the LLM from accessing the full context of the model pool during pipeline generation, further hindering the effectiveness of \LLMPipe.

In addition, \LLMSolve shows strong performance in typical natural language processing tasks such as summarization and translation, 
yet it completely fails on structured perception tasks, like speech recognition and sentence similarity. Particularly, it achieves an average score of up to 0.74 in translation tasks, but yields a score of 0 across all settings for speech recognition tasks. 
Although the employed LLM, \text{GPT-4o-mini}, is a general-purpose LLM with multi-modal capabilities, it remains limited in specialized tasks. In particular, it is not explicitly fine-tuned for automatic speech recognition. Successfully solving these tasks often requires integrating multiple models and incorporating pre/post-processing steps, which are not equipped in \LLMSolve to handle under a direct prompting strategy.
As a result, LLM-based approaches' performance on such tasks remains poor.

The manually constructed \hummanRef outperforms both LLM-based strategies across all task types, with average performance that is 90\% higher than \LLMSolve and 6.7 times higher than \LLMPipe. 
The gap is particularly striking in tasks like speech recognition and sentence similarity, where \hummanRef achieves an average score of 0.95, while both \LLMSolve and \LLMPipe performed extremely poorly on these tasks. \textit{These findings highlight the significant performance gap between human-designed and automated CA solutions and emphasize the need for more capable and robust automatic composition methods.}

\subsubsection{Performance of Baselines across Solution Complexities}



\begin{table}[]
\caption{Performance of baselines across solution complexities}
\label{tab:result_by_pipeline_complexity}
\resizebox{\columnwidth}{!}{%
\begin{tabular}{ll|c|c|c|c}
\toprule
\multicolumn{2}{c|}{\textbf{Approach}}                                        & \multicolumn{1}{l|}{\begin{tabular}[c]{@{}c@{}}\textbf{Atomic} \\ \textbf{solution}\end{tabular}} & \multicolumn{1}{l|}{\begin{tabular}[c]{@{}c@{}}\textbf{Chain} \\ \textbf{solution}\end{tabular}} & \multicolumn{1}{l|}{\begin{tabular}[c]{@{}c@{}}\textbf{Graph} \\ \textbf{solution}\end{tabular}} & \multicolumn{1}{c}{\textbf{Average}} \\ \midrule
\multicolumn{1}{l|}{\multirow{3}{*}{Prompt-to-Solve}}    & Zero-shot & 0.51                                                                        & 0.64                                                                       & 0.38                                                                       & 0.53                         \\  
\multicolumn{1}{l|}{}                                    & Few-shot  & 0.52                                                                        & 0.65                                                                       & 0.40                                                                       & 0.54                         \\ 
\multicolumn{1}{l|}{}                                    & CoT       & 0.49                                                                        & 0.66                                                                       & 0.39                                                                       & 0.52                         \\ \midrule
\multicolumn{1}{l|}{\multirow{3}{*}{Prompt-to-Pipeline}} & Zero-shot & 0.09                                                                        & 0.09                                                                       & 0.07                                                                       & 0.09                         \\  
\multicolumn{1}{l|}{}                                    & Few-shot  & 0.19                                                                       & 0.28                                                                       & 0.10                                                                       & 0.20                         \\ 
\multicolumn{1}{l|}{}                                    & CoT       & 0.20                                                                        & 0.36                                                                       & 0.09                                                                       & 0.23                         \\ \midrule
\multicolumn{2}{l|}{Human-designed reference}                        & 0.82                                                                        & 0.79                                                                       & 0.72                                                                       & 0.80                         \\ \bottomrule
\end{tabular}
}
\end{table}

Table~\ref{tab:result_by_pipeline_complexity} reports the performance of the approaches on tasks categorized by the structural complexity of their corresponding \textit{Human-designed Reference Solutions}.
Overall, \textit{\LLMSolve consistently outperforms \textit{Prompt-to-Pipeline} by a significant margin} across all levels of complexity. Specifically, \LLMSolve achieves an average score of 0.53, which more than three times the performance of \LLMPipe. This demonstrates the effectiveness of directly leveraging the LLM's knowledge for end-to-end task resolution.

However, despite its strong performance, \LLMSolve still lags behind the \hummanRef.
On average, \hummanRef outperforms \LLMSolve by 51\%, with the performance gap is up to 85\% on highly complex tasks such as those requiring graph-structured pipelines. 
This highlights that while LLMs can address these tasks to some extent, specialized models orchestrated through carefully designed pipelines remain substantially more effective, particularly for tasks requiring compositional reasoning and structured execution.
\textit{These results illustrate the promise of composable AI for tackling complex real-world problems and emphasize the need to develop methods for unlocking the full potential of composable AI systems by automatically generating effective pipelines}.

\section{Related Work}

The rapid advancement of AI in recent years has been largely driven by the emergence of LLMs~\cite{wang2025history, annepaka2025large, dubey2024llama}. These models, such as GPT-3, GPT-4, PaLM, and LLaMA, have demonstrated remarkable generalization capabilities across a wide range of tasks, from text generation~\cite{qu2023layoutllm} and summarization~\cite{zhang2025systematic} to mathematical reasoning~\cite{sheng2025learning}, commonsense inference~\cite{dong2025open}, and code generation~\cite{Rambo}.
However, they still often struggle with specialized tasks where domain expertise is required. To bridge this gap, Liang \etal~\cite{liang2024taskmatrix} introduce TaskMatrix.AI which is an  AI ecosystem that connects LLMs with millions of APIs. Given a user request, TaskMatrix.AI analyzes the task and invokes suitable APIs to fulfill it. Similarly, Toolformer~\cite{schick2023toolformer} teaches LLMs to autonomously decide when and how to call external tools via APIs to enhance their performance on complex tasks. 
Different from these approaches, \tool not only focuses on selecting and invoking correct sequences of APIs, but also emphasizes the composition of these components into coherent and executable pipelines. This includes managing interoperability, handling data transformations through glue code, and ensuring that the composed system can solve the original task end-to-end.

Our work is also closely related to HuggingGPT~\cite{shen2023hugginggpt}, which leverages LLMs as a controller to interpret user instructions, plan solutions, and delegate sub-tasks to pre-trained models hosted on the Hugging Face platform. While HuggingGPT explores task decomposition and model selection, it does not fully address the challenge of integrating selected models into executable pipelines. In contrast, \tool provides a benchmark that explicitly evaluates both the correctness of task decomposition and the feasibility of executing composed solutions, highlighting practical constraints such as model compatibility and execution consistency.


\section{Conclusion}

In this work, we take an initial step toward advancing the study of composable AI, a promising paradigm for solving complex tasks by decomposing them into manageable sub-tasks and leveraging existing well-trained models. To enable rigorous and reproducible research in this domain, we introduce \tool, the first public benchmark designed specifically for composable AI. We also develop a full-pipeline evaluation framework that supports end-to-end performance assessment. 
Our experiments with LLM-based baselines highlight both the potential and current limitations of automatic methods in tackling composable AI tasks, especially for tasks requiring compositional reasoning and structured execution.



\bibliographystyle{IEEEtran}

\bibliography{ref}

\end{document}